\documentclass{vgtc}                          

\ifpdf
  \pdfoutput=1\relax                   
  \usepackage{graphicx}                
  \DeclareGraphicsExtensions{.pdf,.png,.jpg,.jpeg} 
\else
  \usepackage{graphicx}                
  \DeclareGraphicsExtensions{.eps}     
\fi%

\graphicspath{{figures/}{pictures/}{images/}{./}} 

\usepackage{microtype}                 
\PassOptionsToPackage{warn}{textcomp}  
\usepackage{textcomp}                  
\usepackage{mathptmx}                  
\usepackage{times}                     
\usepackage{cite}                      
\usepackage{tabu}                      
\usepackage{booktabs}                  

\usepackage{xcolor}

\usepackage{amssymb}
\usepackage{picinpar,moresize,xfrac,nicefrac,graphpap,dcolumn,wrapfig,graphicx,hyperref}
\microtypesetup{stretch=30,shrink=30}

\usepackage{tikz,pgfplots}
\usepgfplotslibrary{colormaps}
\usetikzlibrary{pgfplots.colormaps}
\usepackage{rotating}
\pgfplotsset{compat=newest} 
\pgfplotsset{plot coordinates/math parser=false}
\newlength\figureheight 
\newlength\figurewidth

\usepackage{algorithm}
\usepackage{algorithmicx}
\usepackage{algpseudocode}

\hypersetup{%
  breaklinks,
  colorlinks=true,
  linkcolor=black,
  anchorcolor=black,
  citecolor=black,
  filecolor=black,
  menucolor=black,
  urlcolor=black,
  bookmarksnumbered,
  pdfencoding=auto,
  pdfstartview=FitH,
  pdfpagemode=UseNone,
  pdfnewwindow=true,
  bookmarksopen=false,
  baseurl={}
}

\usepackage{soul,array,calc,url,ragged2e,graphpap}
\usepackage{booktabs} 

\usepackage[figure]{hypcap}
\usepackage{mathtools}
\mathtoolsset{centercolon}

\usepackage[normalem]{ulem}
\usepackage{cancel}

\makeatletter
\DeclareFontFamily{OMX}{MnSymbolE}{}
\DeclareSymbolFont{MnLargeSymbols}{OMX}{MnSymbolE}{m}{n}
\SetSymbolFont{MnLargeSymbols}{bold}{OMX}{MnSymbolE}{b}{n}
\DeclareFontShape{OMX}{MnSymbolE}{m}{n}{
    <-6>  MnSymbolE5
   <6-7>  MnSymbolE6
   <7-8>  MnSymbolE7
   <8-9>  MnSymbolE8
   <9-10> MnSymbolE9
  <10-12> MnSymbolE10
  <12->   MnSymbolE12
}{}
\DeclareFontShape{OMX}{MnSymbolE}{b}{n}{
    <-6>  MnSymbolE-Bold5
   <6-7>  MnSymbolE-Bold6
   <7-8>  MnSymbolE-Bold7
   <8-9>  MnSymbolE-Bold8
   <9-10> MnSymbolE-Bold9
  <10-12> MnSymbolE-Bold10
  <12->   MnSymbolE-Bold12
}{}

\let\llangle\@undefined
\let\rrangle\@undefined
\DeclareMathDelimiter{\llangle}{\mathopen}%
                     {MnLargeSymbols}{'164}{MnLargeSymbols}{'164}
\DeclareMathDelimiter{\rrangle}{\mathclose}%
                     {MnLargeSymbols}{'171}{MnLargeSymbols}{'171}
\makeatother

\newcommand{\secref}[1]{Sec.~\ref{#1}}

\renewcommand{\algref}[1]{Alg.~\ref{#1}}

\DeclareSymbolFont{bbold}{U}{bbold}{m}{n}
\DeclareSymbolFontAlphabet{\mathbbold}{bbold}

\usepackage{xfrac,nicefrac}

\def\CC{\mathbb{C}}

\def\bA{\mathbf{A}}
\def\bI{\mathbf{I}}
\def\bL{\mathbf{L}}
\def\bQ{\mathbf{Q}}

\def\bc{\mathbf{c}}

\def\bp{\mathbf{p}}
\def\bu{\mathbf{u}}
\def\bv{\mathbf{v}}

\def\bp{\mathbf{p}}

\onlineid{5}

\vgtccategory{Research}

\vgtcinsertpkg

\title{Center of circle after perspective transformation\thanks{Preliminary work.}}

\author{Xi Wang\thanks{e-mail:xi.wang@tu-berlin.de}\\ %
        \scriptsize TU Berlin %
\and Albert Chern\thanks{e-mail:chern@math.tu-berlin.de}\\ %
     \scriptsize TU Berlin
\and Marc Alexa\thanks{e-mail:marc.alexa@tu-berlin.de}\\ %
     \scriptsize TU Berlin
     }

\abstract{
Video-based glint-free eye tracking commonly estimates gaze direction based on the pupil center. The boundary of the pupil is fitted with an ellipse and the euclidean center of the ellipse in the image is taken as the center of the pupil.
However, the center of the pupil is generally not mapped to the center of the ellipse by the projective camera transformation.
This error resulting from using a point that is not the true center of the pupil directly affects eye tracking accuracy.
We investigate the underlying geometric problem of determining the center of a circular object based on its projective image. The main idea is to exploit two concentric circles -- in the application scenario these are the pupil and the iris. We show that it is possible to computed the center and the ratio of the radii from the mapped concentric circles with a direct method that is fast and robust in practice.
We evaluate our method on synthetically generated data and find that it improves systematically over using the center of the fitted ellipse. Apart from applications of eye tracking we estimate that our approach will be useful in other tracking applications.

} 

\CCScatlist{ 
\CCScat{}{Computing Methodologies}{Computer Vision}{Model development and analysis}
}

\begin{document}


\firstsection{Introduction}

\maketitle



It is becoming more and more common to include video cameras in head mounted stereo displays to enable eye tracking in virtual reality environments. In this setting, the optical axis of the eye is usually determined based on the center of the pupil, which can be extracted from the video images. Identification of the pupil in the image stream is commonly done by fitting an ellipse to the boundary between the dark pupil and much lighter iris. This approach introduces two sources of error: 1) the center of the ellipse in the image is not the center of the pupil, because the camera transformation is projective (and not just affine); 2)  the refraction at the cornea distorts the pupil shape in addition to the projective camera transformation. The problems from refraction can be circumvented in other situations by additional tracking equipment~\cite{duchowski2007eye}. Without such equipment it may be possible to compensate the effect based on a computationally involved inverse model~\cite{Dierkes2018}. We focus on the first problem, namely the non-affine mapping of the center. To our knowledge, this is the first work 
providing a computational approach estimating the true pupil center in the context of video-based eye tracking. 

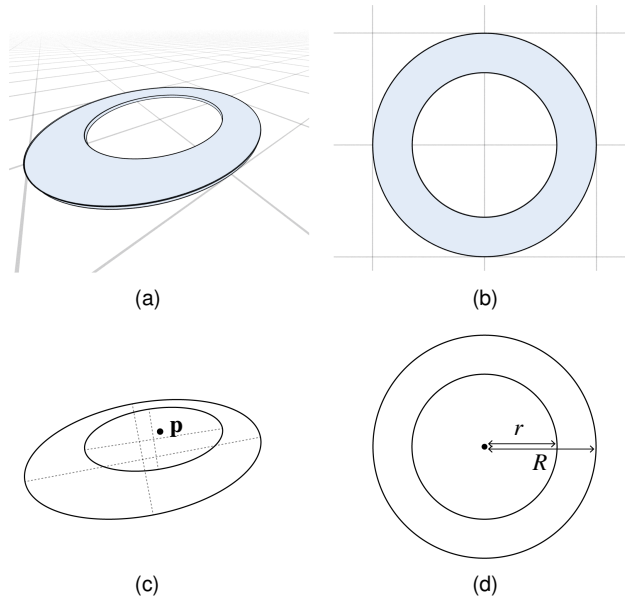
\begin{figure}[t]
    \centering
    \setlength{\unitlength}{0.01\columnwidth}
    \begin{picture}(100,89)(0,0)
        \put(0,-1){\includegraphics[width=\columnwidth]{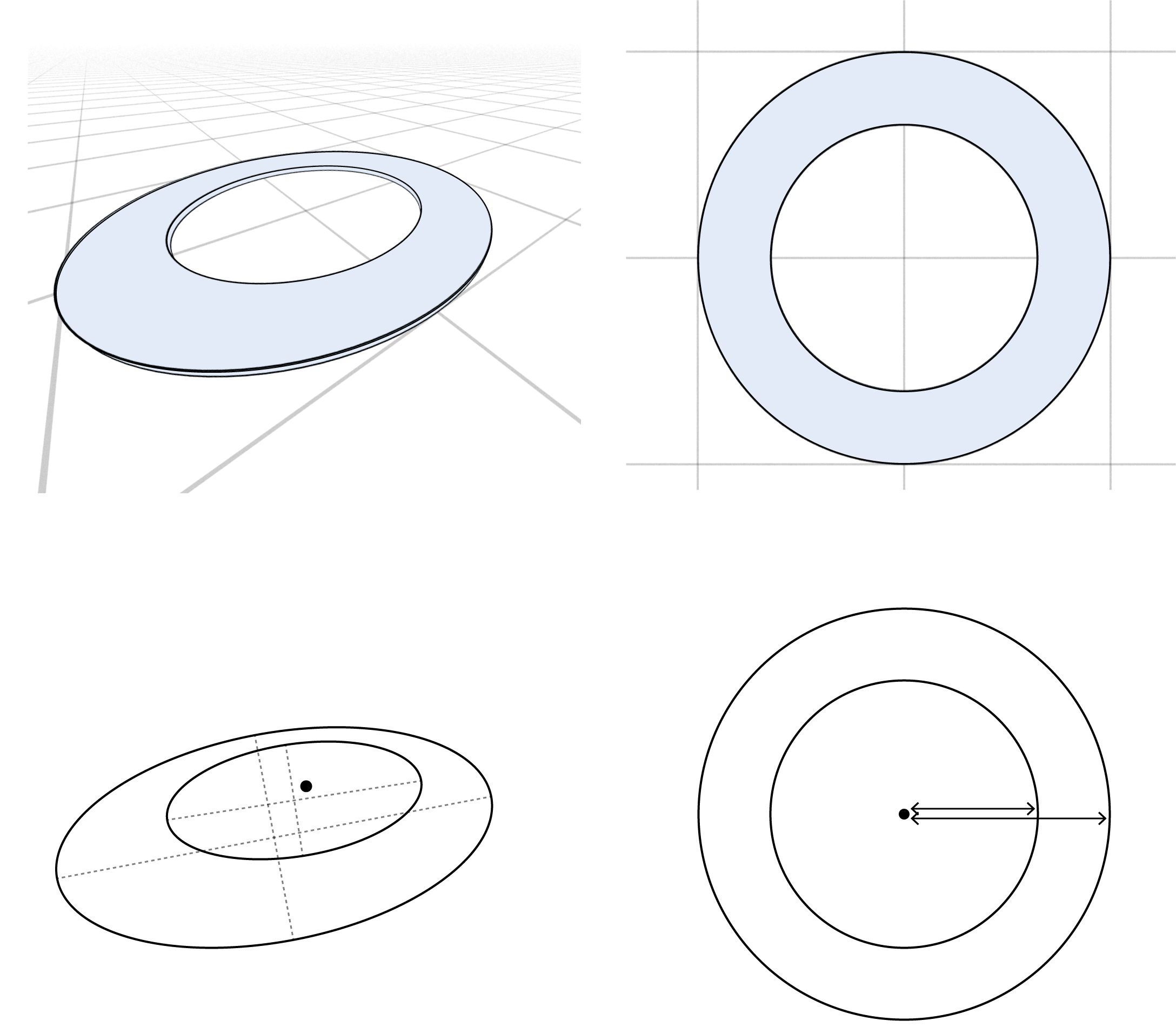}}
        \put(22,40){\textsf{\scriptsize(a)}}
        \put(75,40){\textsf{\scriptsize(b)}}
        \put(22,-5){\textsf{\scriptsize(c)}}
        \put(75,-5){\textsf{\scriptsize(d)}}
        \put(81.5,19.5){\(r\)}
        \put(84.5,14){\(R\)}
        \put(27.5,20){\(\bp\)}
    \end{picture}
    \caption{\label{fig:Teaser}
    A pair of ellipses (a) as the image of a pair of concentric circles (b) in perspective.  Taking the data from a conventional ellipse detection tool (shown in axes in (c)), a simple algorithm proposed in this paper locates the true center \(\bp\) respecting the perspective, and determines the radii ratio \(\nicefrac{R}{r}\) of the inferred circles (d).
    }
\end{figure}

The fact that the center of a circle is not mapped to the center of an ellipse under projective transformations is illustrated in Figure~\ref{fig:Teaser}: The pair of ellipses in (a) is the result of perspectively projecting the two concentric circles in (b) to the image plane. The centers of the two ellipses are not coincident (c), and neither coincides with the projected center $\bp$ of the concentric circles. 
As we explain later, there are in fact many projective transformations that would map a pair of concentric circles to the two ellipses found in the image, however, all of them give rise to the \emph{same} projective center, and also all of them agree on the ratio of the radii. In other words, based on the pair of ellipses, the center of the concentric circles and the ratio of the radii are uniquely determined.

The set of concentric circles is an instance of a \emph{pencil of conics}. One of the degenerated circles in the group corresponds to the true center and it is invariant under projective transformation. In Section~\ref{sec:MathematicalFoundation} we explain this concept and show how  it leads to a simple formulation for a computational approach.

We apply our method to estimate the true pupil center by using a pair of ellipses, i.e., the pupil ellipse and the iris ellipse. Tested with synthetic data, we show that our method provides robust estimation of pupil center with respect to projective distortion (see Section~\ref{sec:results}). Compared to the pupil center estimated from ellipse fitting our estimation shows significant improvement, less than one pixel distance to the true pupil center in most cases. 

Apart from eye tracking, our method can also be used for related tasks in computer vision. The idea of exploring the projective invariant properties of concentric circles is not new. It has been used for markers consisting of the two concentric circles specified by a ring ~\cite{1541275} or localization in robotics applications~\cite{6630694}. Compared to the iterative optimization techniques our formulation as an eigenproblem is more direct.

\section{Background and Related Work}


The estimation of gaze direction in video-based eye tracking relies on measuring the pupil center~\cite{duchowski2007eye, holmqvist2011eye}.  Image analysis algorithms are used to estimate the pupil center. The boundary of the pupil is fitted into an ellipse, and its euclidean center is taken as the pupil center~\cite{Swirski2012, Fuhl2016}. The estimated pupil center is then mapped to gaze direction through calibration, very often with the aid of glint, known as the corner reflection, reflected from a controlled light source. However, most mobile eye tracking system are based on glint-free tracking systems~\cite{Kassner2014}.

How to accurately estimate pupil center position is of major concern in the eye tracking community. The main challenges are risen from distortions in the captured images, namely the projective distortion and the refractive distortion~\cite{4564194, Dierkes2018}. Projective distortion moves the projected pupil center away from the ellipse center as shown in Figure~\ref{fig:Teaser}. The refractive distortion is caused by the different refractive index of cornea ($n=1.33$, $n_{air}=1.0$~\cite{atchison2000optics}), which leads to irregularly distorted pupil boundaries in the camera image. Such distortion depends on the camera viewing direction~\cite{Fedtke10} even when the cornea surface is simplified to a perfect sphere. The analysis gets more complicated when the curvature of the cornea changes~\cite{barsingerhorn2017optics}. Glint-based eye tracking systems implicitly model the refraction distortion~\cite{1634506}, for example by using the difference vector between pupil center and corner reflection for gaze estimation. 

Apart from the above described feature-based gaze estimation, appearance-based gaze estimation methods rely on the detection of eye features in the images~\cite{Mora2014, Park2018}, such as the iris center and the eye corners. These methods aim to track the eye movements for example with webcams. Image resolution in the eye region is limited and the appearances of pupil and cornea are less distinguishable. Very often full faces are visible in the images, therefore, facial landmarks are used as additional information in deep learning methods~\cite{Palmero2018RecurrentCF, Zhang2019}. However, simultaneous detection of both pupil and iris is difficult in both feature-based methods and appearance-based methods, and it inevitably requires many empirical parameter settings~\cite{Yeong2013, szczepanski2014pupil}. Robust detection of both pupil and iris remains as a challenging problem, especially that irises are partially occluded by eyelids. 

Additionally, model-based gaze estimations have been proposed where multiple eye images are used to optimize a three-dimensional eye ball model~\cite{Swirski2013}. Recently a model-fitting approach accounting for cornea refraction has been proposed~\cite{Dierkes2018}. In principle, these model fitting methods are based on ray tracing, which is a considerably expensive procedure. It results in a non-linear minimization problem which requires iterative solving procedures.  

We propose a simple method utilizing the underlying geometric properties of two concentric circles, the pupil and the iris. Our method directly computes the true pupil center in the camera image as well as the projective invariant radii ratio. In general, large distance to the camera image center leads to large projective distortion. In the context of eye tracking, this means larger pupils are more seriously distorted. Recent study~\cite{HOOGE20191} shows that changes in pupil sizes can lead to sever accuracy drops in eye tracking. Pupil size has been used to study observer's mental cognitive load~\cite{Palinko2010} as well as fatigues when using VR headsets~\cite{UKAI2008106}. Our method can also be used to accurately estimate the pupil size (in these applications). 



\section{Concentric Circles in Perspective}
Given an ellipse that is the result of the projective transformation of a circle, is it possible to identify the projected center of the circle?  The answer to this question is no.  There is an infinite set of projective transformations that map a circle to the observed ellipse, while sending the center to an arbitrary points.  Therefore, given only an ellipse from a camera view without any information of the perspective, one cannot retrieve the image of the center of the original  circle.  This center retrieval problem, however, becomes drastically different when the given image contains \emph{two} ellipses coming from a pair of \emph{concentric circles}. In the following we introduce the main observations and the resulting algorithm -- the following section provides the mathematical justification. 

Given two ellipses (or more generally, conics)
there may be many projective transformations that map the ellipses into an inferred pair of concentric circles. Nevertheless, these possible transformations will all agree on
\begin{enumerate}
    \item a unique point \(\bp\) that is sent to the center of the target concentric circles; and
    \item the ratio \(\nicefrac{R}{r}\) of the radius \(R\) of the larger circle to the radius \(r\) of the smaller circle.
\end{enumerate}
In other words, based on only two ellipses in an image, and without any information on the projective transformation from the eye to the image plane of the camera, one can identify the center of the concentric circles in the image and compute the radii ratio of the circles.

It turns out that finding the center point \(\bp\) and the radii ratio only amounts to an eigenvalue problem of a \(3\times 3\) matrix, based on which we give a simple algorithm for the center retrieval problem.

\subsection{Matrix Representation of Conics}
A conic, such as an ellipse, takes a general implicit form
\begin{align}
    \label{eq:ImplicitGeneral}
    Ax^2 + Bxy + Cy^2 + Dx + Ey + F = 0,
\end{align}
which can be expressed in the following matrix form:
\begin{align*}
    \begin{bmatrix}
        x&y&1
    \end{bmatrix}
    \underbrace{\begin{bmatrix}
        A&\nicefrac{B}{2} & \nicefrac{D}{2}\\
        \nicefrac{B}{2} & C & \nicefrac{E}{2}\\
        \nicefrac{D}{2} & \nicefrac{E}{2} & F
    \end{bmatrix}}_{\bQ}
    \begin{bmatrix}
        x\\y\\1
    \end{bmatrix}
    = 0,
\end{align*}
where the symmetric matrix \(\bQ\) is called the matrix representation of the conic, and \(\widetilde\bv = (x,y,1)^\intercal\) is a vector in \emph{homogeneous coordinates}.
Note that the rescaled matrix \(\bQ\mapsto \alpha \bQ\), where \(\alpha\) is a nonzero scalar, defines the same conic. This means the matrix representation is a homogeneous coordinate of the conic. Each conic uniquely corresponds to such a homogeneous matrix representation. 

For an ellipse with geometric parameters \(\bc = (c_x, c_y)\) (geometric centroid), \(a\) (major semiaxis), \(b\) (minor semiaxis) with a rotation angle \(\theta\), the coefficients in \eqref{eq:ImplicitGeneral} are given by
\begin{align*}
    &A = a^2\sin^2\theta + b^2\cos^2\theta,
    &&B = 2(b^2 - a^2)\cos\theta\sin\theta, \\
    &C = a^2\cos^2\theta + b^2\sin^2\theta,
    &&D = -2Ac_x -Bc_y,\\
    &E = -Bc_x -2Cc_y,
    &&F = Ac_x^2 + Bc_xc_y + Cc_y^2 - a^2b^2.
\end{align*}

\subsection{Algorithm}
Let \(\bQ_1, \bQ_2\) be two detected ellipses, represented in matrix form, and assume they are the result of an unknown projective transformation of a pair of concentric circles.  Then the following algorithm finds the center \(\bp\) and the radii ratio \(R/r\) for the inferred pair of concentric circles.
\begin{algorithm}[H]
    \caption{\label{alg:Main}Concentric circles in perspective}
\begin{algorithmic}[1]
\Require \(\bQ_1, \bQ_2\)\Comment Two ellipses as \(3\times 3\) symmetric matrices.
\State\(\bA\coloneqq \bQ_2\bQ_1^{-1}\).
\State\((\lambda_i,\bu_i)_{i=1}^3\gets\) the eigenvalue and eigenvector pairs of \(\bA\), with the observation that \(\lambda_1\approx\lambda_2\) and \(\lambda_3\) distinguished.
\State\(\widetilde\bp = (\widetilde p_x,\widetilde p_y, \widetilde p_z)^\intercal \coloneqq \bQ_1^{-1}\bu_3\).
\State\(\bp \coloneqq (\nicefrac{\widetilde p_x}{\widetilde p_z},\nicefrac{\widetilde p_y}{\widetilde p_z})\).
\Ensure \(\bp\) and \(\nicefrac{R}{r}\coloneqq \sqrt{\nicefrac{\lambda_2\lambda_3}{\lambda_1^2}}\).
\end{algorithmic}
\end{algorithm}

\section{Mathematical Justification}
\label{sec:MathematicalFoundation}
The rather straightforward algorithm \algref{alg:Main} is derived based on projective geometry.  In this section we provide the necessary background about conics in projective planes, and derive the formulae in \algref{alg:Main}.

\subsection{Special Sets of Conics}
In the following we provide a number of basic relations between the algebraic and the geometric aspects of conics.  These notions allow us to characterize projective transforms of concentric circles.
\subsubsection{Degenerate Conics}
A conic, represented as a \(3\times 3\) symmetric matrix \(\bQ\), is \emph{degenerate} if \(\det(\bQ) = 0\).  What this means geometrically is that the conic becomes one point, one line, a union of two lines, or a union of two complex conjugate lines whose intersection is a real point.

\subsubsection{Projective Transformations}
\label{sec:ProjectiveTransformationOfConics}
A projective transformation deforms a conic \(\bQ\) into another conic with matrix \(\bL^{-\intercal}\bQ\bL^{-1}\), where \(\bL\) is some general invertible \(3\times 3\) matrix representing the non-degenerate projective transformation in homogeneous coordinates.

\subsubsection{Intersections of Conics} Two generic conics \(\bQ_1\), \(\bQ_2\) can have four, two, or zero real intersection points.  When the number of intersection points is less than four, what happens is that the missing intersections become imaginary.  If one allows \(x,y\) to be complex, then there are always four intersection points, counted with multiplicity.

\subsubsection{Pencil of Conics}Given two generic conics \(\bQ_1\), \(\bQ_2\), one can construct a family of conics through linear combinations
\begin{align*}
    \bQ_{(\alpha,\beta)} \coloneqq \alpha\bQ_1 +\beta\bQ_2,\quad \alpha,\beta\in\CC.
\end{align*}
This family of conics is called a \emph{pencil of conics}.  Geometrically, the pencil of conics consists of all conics passing through the four fixed (possibly complex) intersection points of \(\bQ_1\) and \(\bQ_2\).  Since a rescaling of the matrix \(\bQ_{(\alpha,\beta)}\) results in the same conic, one may use only one parameter \(\lambda\) replacing \((\alpha,\beta)\) to parametrize the pencil of conics:
\begin{align}
    \label{eq:Pencil}
    \bQ_{(\lambda)}\coloneqq -\lambda\bQ_1 + \bQ_2,\quad\lambda\in\CC.
\end{align}
The minus sign here is for later convenience.

\subsubsection{Circles}A conic \(\bQ\) is a \emph{circle} if \(A = C\) and \(B = 0\) in \eqref{eq:ImplicitGeneral}.  This condition is equivalent to equation \(\widetilde\bv^\intercal\bQ\widetilde\bv = 0\) admitting two special solutions \(\widetilde\bv = (1,i,0)^\intercal\) and \(\widetilde\bv = (1,-i,0)^\intercal\).  The two complex points at infinity  \((1,\pm i, 0)^\intercal\) are called the \emph{circular points}.  (Here, ``point at infinity'' refers to the vanishing 3rd component in the homogeneous coordinate.)

\subsubsection{Concentric Circles}
\label{sec:ConcentricCircles}
 Two conics \(\bQ_1\), \(\bQ_2\) are \emph{concentric circles} if they are not only circles (passing through the circular points \((1,\pm i,0)^\intercal\)) but also that \(\bQ_1\), \(\bQ_2\) intersect \emph{only} at the circular points with multiplicity two.  In other words, \(\bQ_1\), \(\bQ_2\) \emph{touch} at the circular points. The pencil of conics spanned by a pair of concentric circles \(\bQ_1, \bQ_2\) consists of all circles concentric to \(\bQ_1\) and \(\bQ_2\).  In this pencil of concentric circles there are several degenerate conics.  One of them corresponds to the circle collapsed to the center point.  This center point is a degenerate conic as the two complex conjugate lines joining the center point to each circular points.  Another degenerate concentric circle is the single real line connecting the two circular points.

\subsection{Projective Transforms of Concentric Circles}
From \secref{sec:ConcentricCircles} we conclude that two real conics \(\bQ_1\), \(\bQ_2\) can be projectively transformed into a pair of concentric circles if and only if \(\bQ_1\) and \(\bQ_2\) touch at two complex points.  

The statement follows from that under any real projective transformation, the touching points \((1,\pm i,0)^\intercal\) of any pair of concentric circles are transformed to some other pair of complex conjugated points, and at the same time the incidence relations---such as the notion of touching---are preserved.  Conversely, if two real conics touch at two complex points, then these two touching points must be the complex conjugate of each other (since the conics are real).  Hence there exists real projective transformations sending the two touching points to \((1,\pm i,0)^\intercal\).  Such projective transformations effectively map the two conics into a pair of concentric circles. 

\subsubsection{Finding the Center}
Suppose \(\bQ_1\) and \(\bQ_2\) are projective transforms of a pair of concentric circles.  Then finding their common center amounts to seeking a degenerate conic in the pencil of conics spanned by \(\bQ_1\) and \(\bQ_2\).  Using the parametric equation \eqref{eq:Pencil} for the pencil, we solve
\begin{align}
    \label{eq:PreEigenvalueProblem}
    \det\left(-\lambda\bQ_1 + \bQ_2\right) = 0.
\end{align}
Assuming \(\det(\bQ_1)\neq 0\) we rewrite \eqref{eq:PreEigenvalueProblem} as
\begin{align*}
    \det\left(-\lambda\bI + \bQ_2\bQ_1^{-1}\right) = 0,
\end{align*}
which is an eigenvalue problem for
\begin{align*}
    \bA\coloneqq\bQ_2\bQ_1^{-1}.
\end{align*}
The three eigenvalues \(\lambda_1,\lambda_2,\lambda_3\) correspond to three degenerate conics in the pencil.  Two of the degenerate conics coincide (\(\lambda_1=\lambda_2\)), being the the single line joining the two touching points of \(\bQ_1,\bQ_2\). The other distinguished degenerate conic is a real point \(\bp\) (together with two complex conjugate lines) representing the center we look for.

The point \(\bp\) in the degenerate conic \(\bQ_{(\lambda_3)} = -\lambda_3\bQ_1+\bQ_2\) can be found by solving \(\bQ_{(\lambda_3)}\bp = 0\):
\begin{align*}
   & (-\lambda_3\bQ_1+\bQ_2)\bp = 0\quad\implies\quad
   \bA\bQ_1\bp = \lambda_3\bQ_1\bp.
\end{align*}
That is, \(\bp = \bQ_1^{-1}\bu_3\) where \(\bu_3\) is the eigenvector \(\bA\bu_3 =\lambda_3\bu_3\) associated with the eigenvalue \(\lambda_3\).

\subsubsection{Radii Ratio}
A projective transformation (\secref{sec:ProjectiveTransformationOfConics}) \(\bQ_1\mapsto\bL^{-\intercal}\bQ_1\bL^{-1}\), \(\bQ_2\mapsto\bL^{-\intercal}\bQ_2\bL^{-1}\) yields
\begin{align*}
    \bA{}\mapsto{}& (\bL^{-\intercal}\bQ_2\bL^{-1})(\bL^{-\intercal}\bQ_1\bL^{-1})^{-1}\\ 
    &= \bL^{-\intercal} \bQ_2\bQ_1^{-1}\bL^{\intercal} = \bL^{-\intercal}\bA\bL^{\intercal},
\end{align*}
which leaves the eigenvalues of \(\bA\) invariant.  Therefore, the ratios  of eigenvalues \(\lambda_1\colon\lambda_2\colon\lambda_3\) are invariant quantities under projective transformations.  Here we only consider the ratios since the matrices \(\bQ_1\), \(\bQ_2\), \(\bA\) are defined only up to a scale.

For \(\bQ_1\), \(\bQ_2\) that are projective transforms of concentric circles, consider a projective transformation \(\bQ_1' = \bL^{-\intercal}\bQ_1\bL^{-1}\), \(\bQ_2' = \bL^{-\intercal}\bQ_2\bL^{-1}\) so that \(\bQ_1'\) and \(\bQ_2'\) are concentric circles with radii \(r\) and \(R\) respectively.  Since the radii are invariant under translations, we may assume that the concentric circles are centered at the origin without loss of generality.  Then we have
\begin{align*}
    \bQ_1' = 
    \begin{bmatrix}
        1&0&0\\ 0&1&0\\ 0&0&-r^2
    \end{bmatrix},\quad
    \bQ_2' = 
    \begin{bmatrix}
        1&0&0\\ 0&1&0\\ 0&0&-R^2
    \end{bmatrix}
\end{align*}
and
\begin{align*}
    \bA' = \bL^{-\intercal}\bA\bL^{\intercal} = \bQ_2' \bQ_1^{\prime -1} = 
    \begin{bmatrix}
        1 & 0 & 0\\ 0&1&0\\0&0& \nicefrac{R^2}{r^2}
    \end{bmatrix}.
\end{align*}
Thus the invariant eigenvalue ratio of \(\bA\) is given by \(\lambda_1\colon\lambda_2\colon\lambda_3 = 1\colon 1\colon\nicefrac{R^2}{r^2}\).  Therefore, the radii ratio \(\nicefrac{R}{r}\) is encoded in the eigenvalues \(\lambda_1 = \lambda_2, \lambda_3\) of \(\bA\).  

In practice, when the two conics \(\bQ_1, \bQ_2\) are detected with measurement error, the spectrum of \(\bA\) may only have an approximated double \(\lambda_1\approx\lambda_2,\lambda_3\).  In that case we retrieve the radii ratios of the concentric circles by one of the following symmetrizations
\begin{align*}
    \frac{R}{r} \approx \sqrt{\frac{\lambda_2\lambda_3}{\lambda_1^2}}
    \approx \sqrt{\frac{\lambda_1\lambda_3}{\lambda_2^2}} \approx\sqrt{\frac{\lambda_3^2}{\lambda_1\lambda_2}}.
\end{align*}


\section{Results}
\label{sec:results}

We apply our method to estimate the true pupil center using the pupil ellipse and the iris ellipse and evaluate it on synthetic data. 

\subsection{Experimental Setup}

Synthetic eye images are rendered using the 3D model proposed in~\cite{Swirski2014} with cornea refractive index set to be 1.0 (see Figure~\ref{fig:ps} for examples). The true projected pupil center in the camera image is directly computed from the 3D model, and we also compute the pupil ellipse and the iris ellipse in the image plane. The pupil center estimated by our method is compared to 1) the Euclidean center of the pupil ellipse in the image plane and 2) the pupil center estimated by image based ellipse fitting following the method proposed in~\cite{Swirski2012}.

Eye camera is placed $3$ cm in front of the eye, similar to the camera position in a mobile eye tracker. We experiment with various viewing angles (rotations of the camera), and three different pupil sizes when the eyes fixate at various targets. Fixation targets are evenly sampled from a circle that is placed perpendicular to the ground. We compare the Euclidean distance in pixel unit between the true pupil center in the image plane and the estimated pupil center using different methods. 

\subsection{Camera Rotation}

With a fixed distance to the eye, the camera is placed at a set of locations that are possible in practice. We use spherical coordinates to describe the rotations. Polar angle $\phi$ defines the rotation angle between the camera and the horizontal plane, and azimuthal angle $\theta$ describes the camera rotation in the horizontal plane. 

\begin{figure}[!t]
    \setlength{\unitlength}{0.01\columnwidth}
    \begin{picture}(100,70)(0,0)
        \put(0,38){\includegraphics[width=.49\columnwidth]{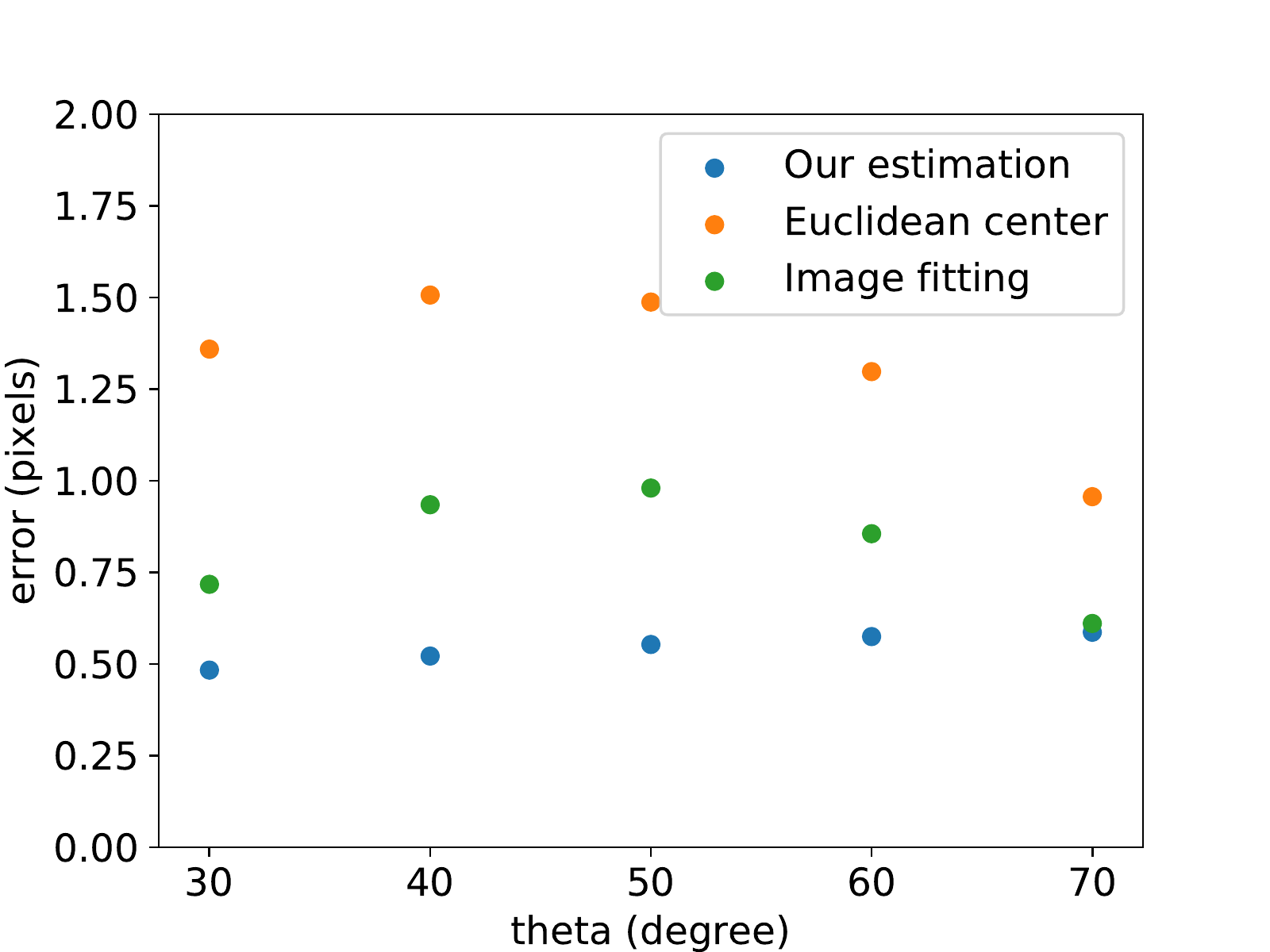}}
        \put(21,35){\textsf{\scriptsize(a)}}
        \put(50,38){\includegraphics[width=.49\columnwidth]{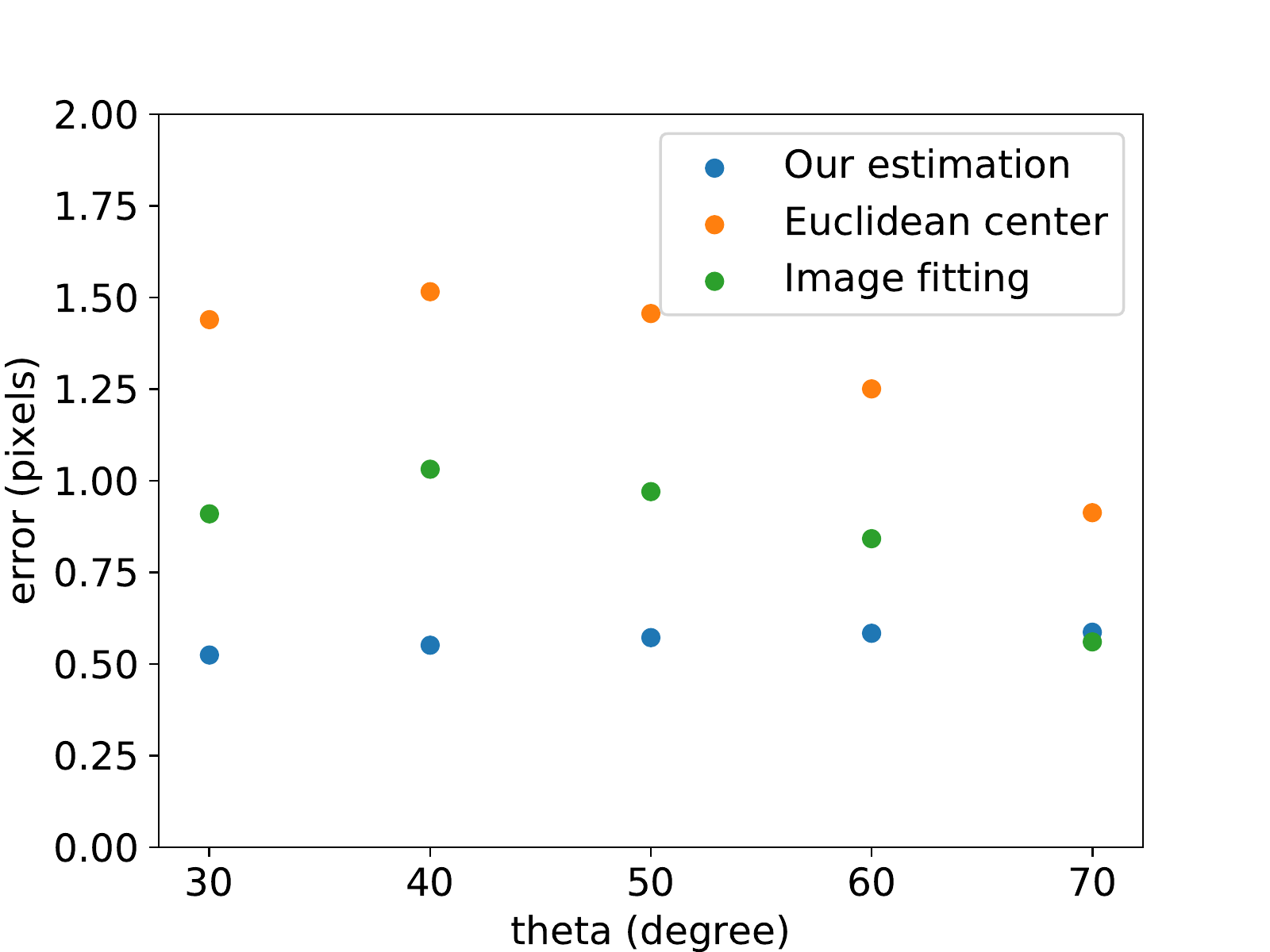}}
        \put(71,35){\textsf{\scriptsize(b)}}
        \put(0,-2){\includegraphics[width=.49\columnwidth]{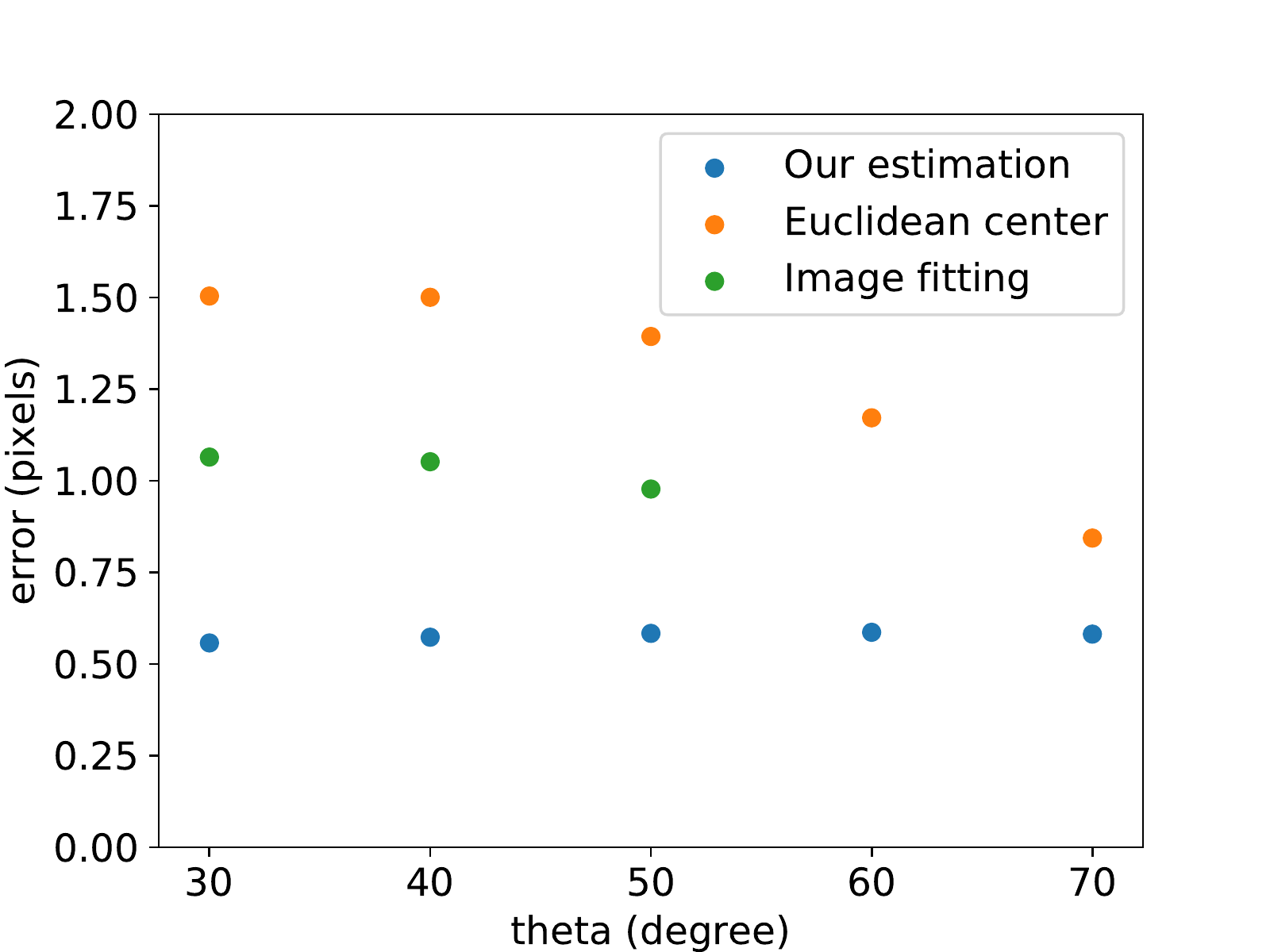}}
        \put(22,-5){\textsf{\scriptsize(c)}}
        \put(50,-2){\includegraphics[width=.49\columnwidth]{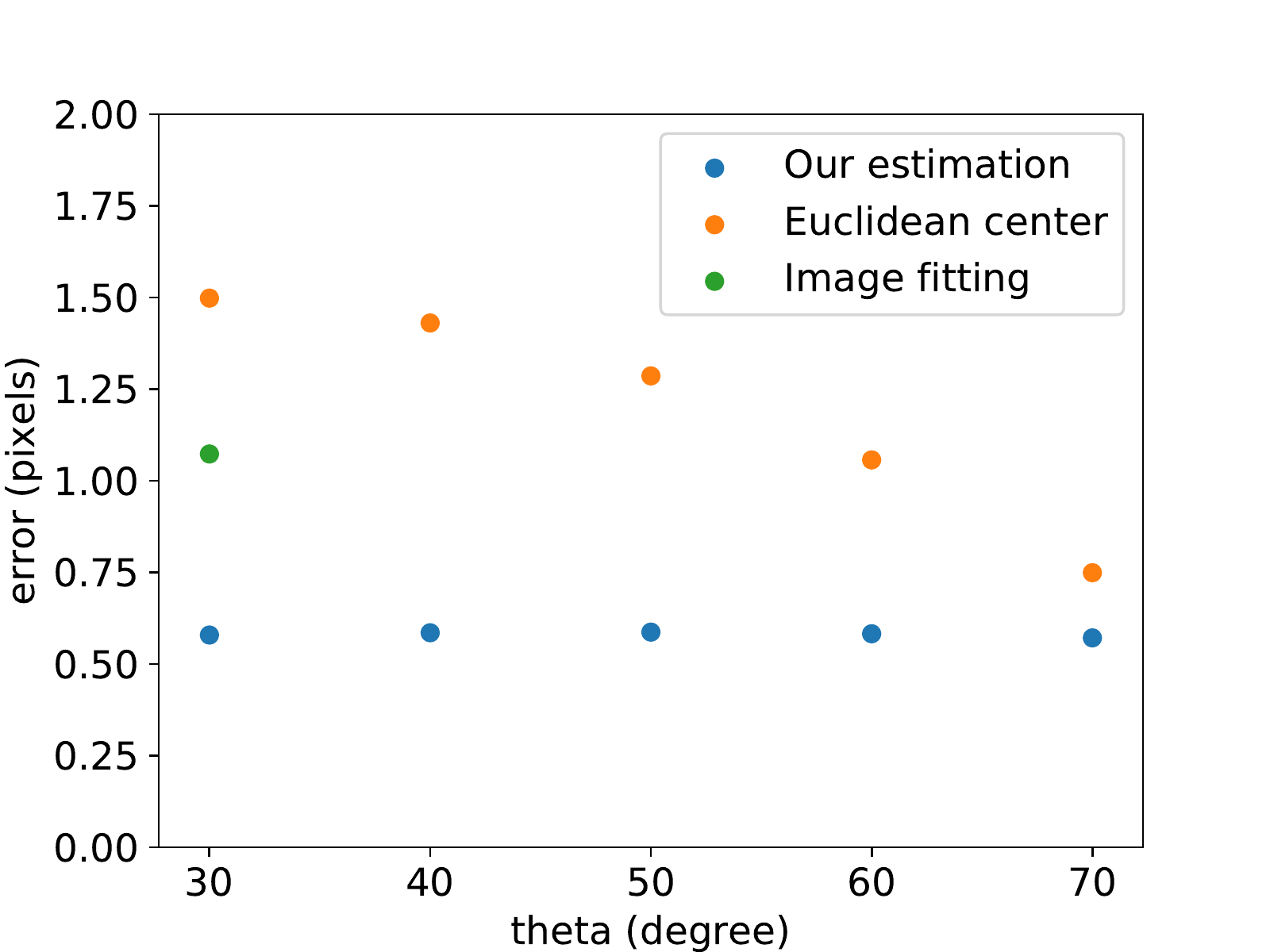}}
        \put(71,-5){\textsf{\scriptsize(d)}}
    \end{picture}
\caption{Estimation errors in different camera positions. $x$ axis corresponds to rotation angle $\theta$ in degree and $y$ axis is the estimation error measured in pixels. From (a) to (d), rotation angle $\phi$ changes from $10^\circ$ to $40^\circ$. In each plot, $\theta$ varies from $30^\circ$ to $70^\circ$.}
\label{fig:cam}
\end{figure}
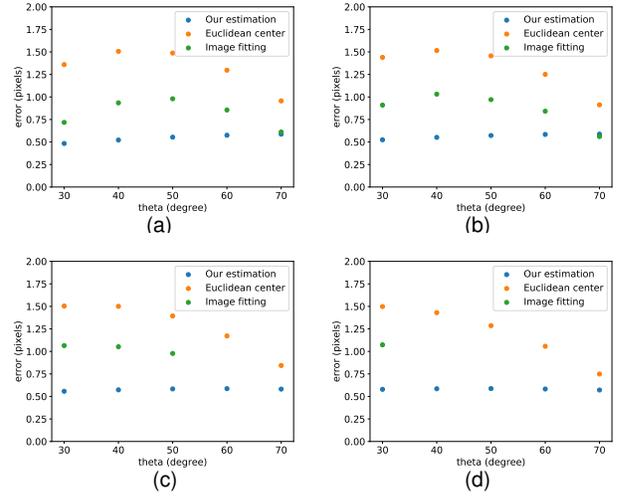

Figure~\ref{fig:cam} shows the estimation errors using different methods. We test with $\phi$ varying from $10^\circ$ to $40^\circ$ and $\theta$ varies from $30^\circ$ to $70^\circ$. In all tested scenarios, our method gives the best estimation with less than one pixel distance to the true pupil center. Estimated Euclidean centers of the pupil ellipse deviate away from the true pupil center. Image based ellipse fitting gives better estimation as more sample points from the boundary detection are used for ellipse fitting. However, the fitting fails when $\theta$ is large. In such cases, the projected pupil is small in the camera image, subsequently with less camera distortion. Therefore, the estimated pupil positions get closer to its true position. As shown in Figure~\ref{fig:cam}, estimation errors decreases with an increasing $\theta$ from left to right in each plot. 

\subsection{Pupil Size}

\begin{figure*}[!t]
    \setlength{\unitlength}{0.01\columnwidth}
    \begin{picture}(100,44)(0,0)
	 \put(0,-3){\includegraphics[width=.33\textwidth]{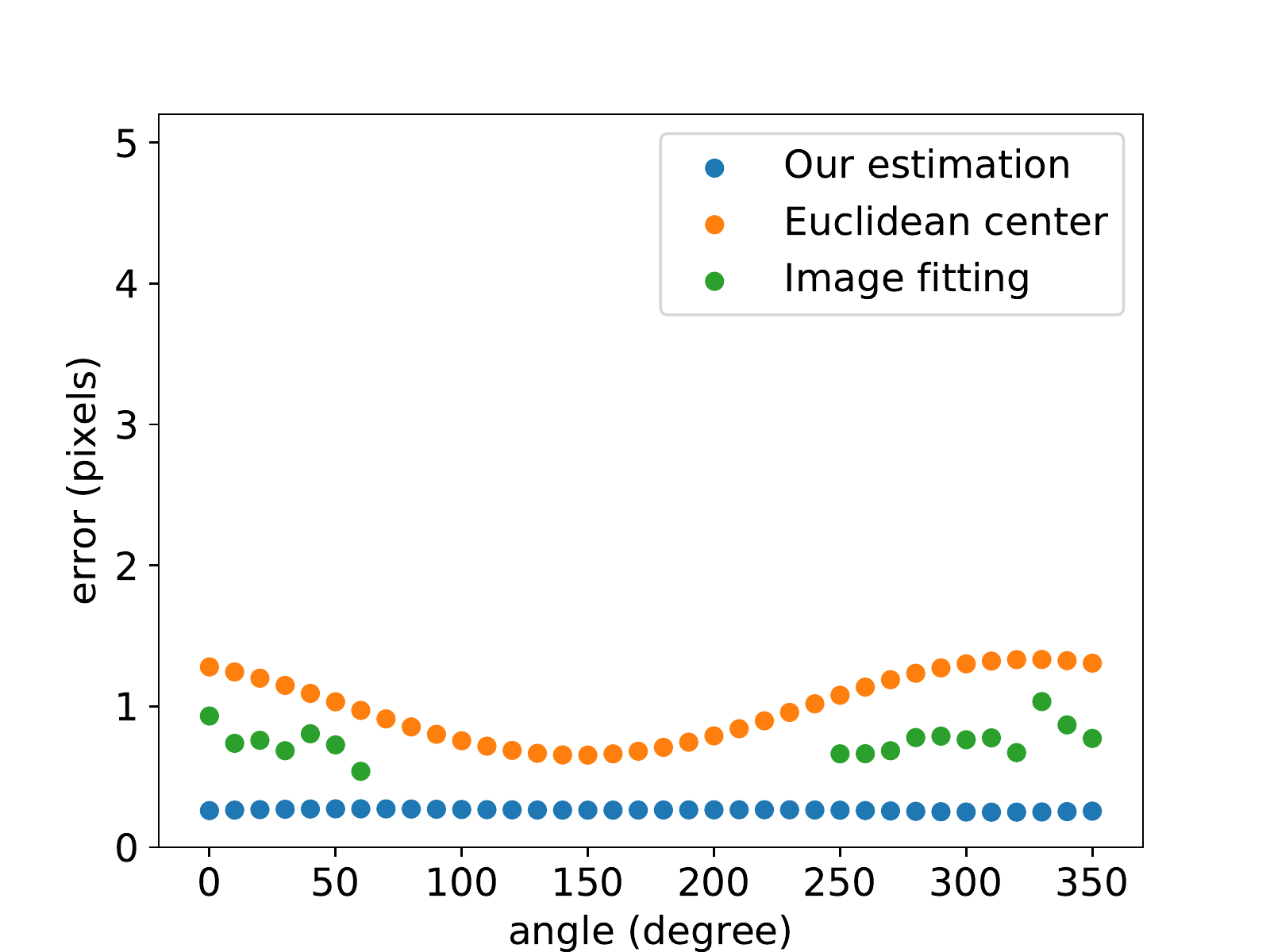}}
	 \put(10, 31){\includegraphics[width=0.15\columnwidth]{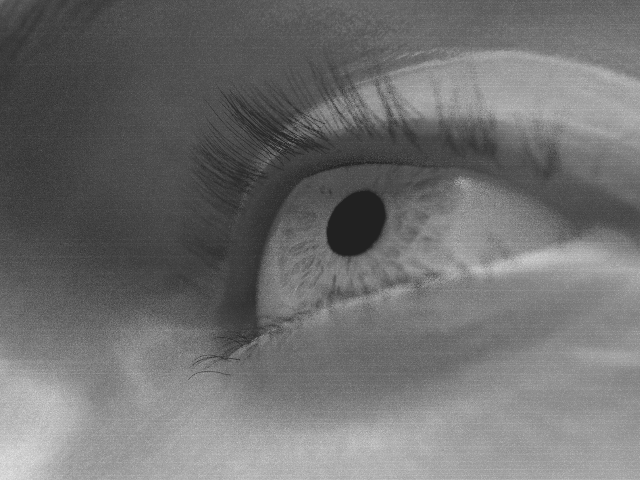}}
	 \put(25, 21){\includegraphics[width=0.15\columnwidth]{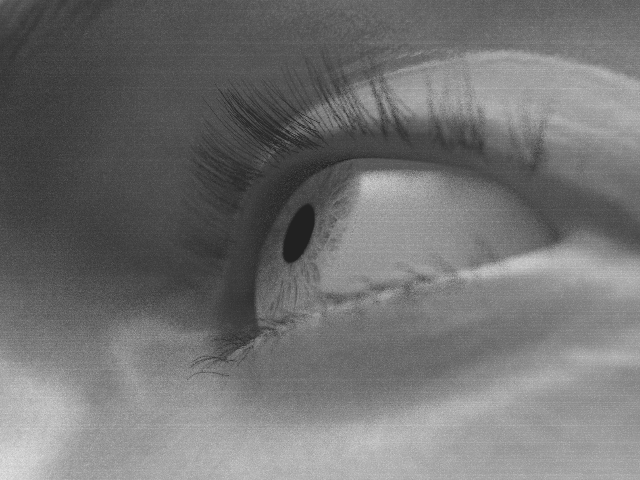}}
	 \put(70,-3){\includegraphics[width=.33\textwidth]{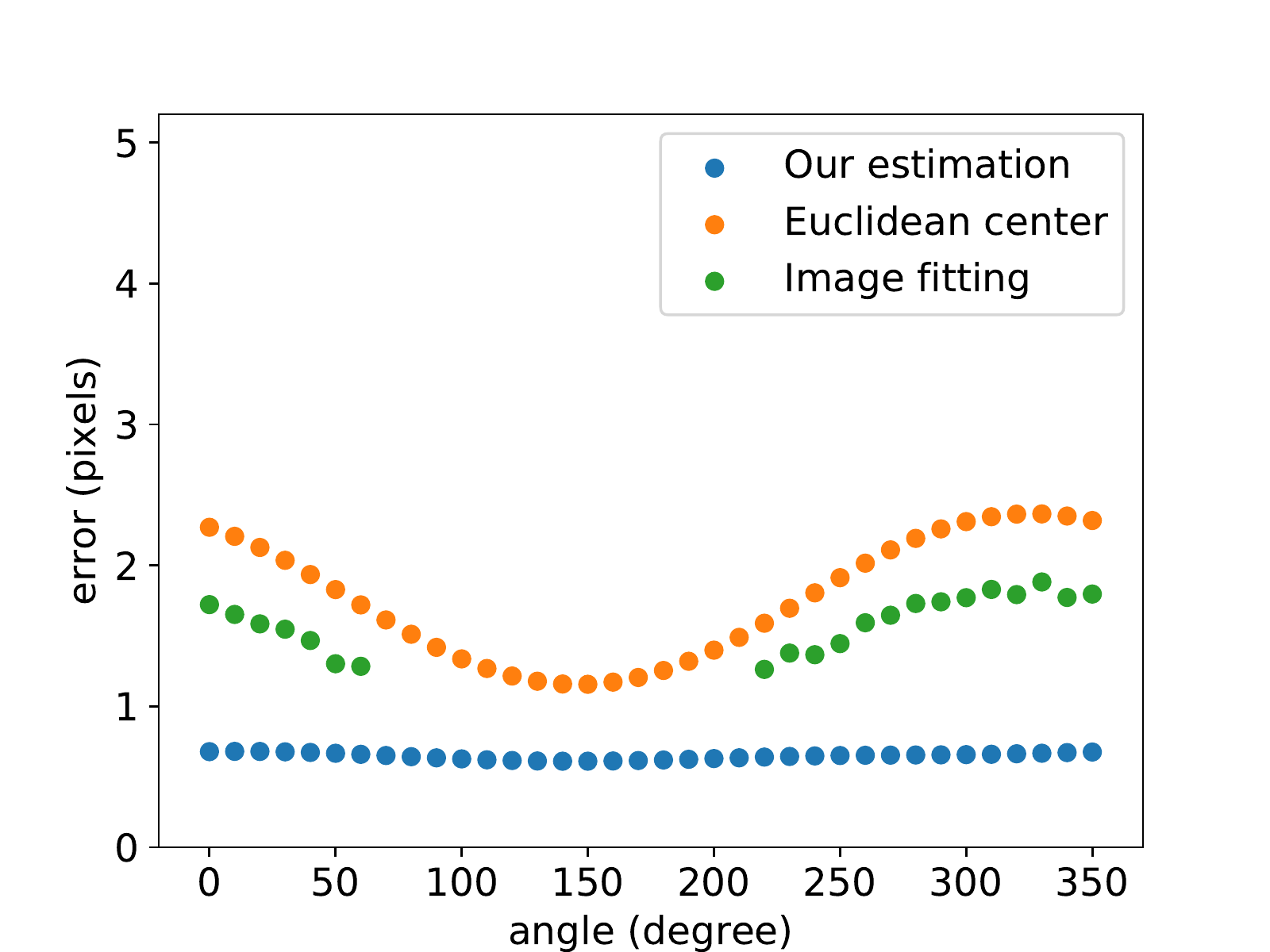}}
	 \put(80, 31){\includegraphics[width=0.15\columnwidth]{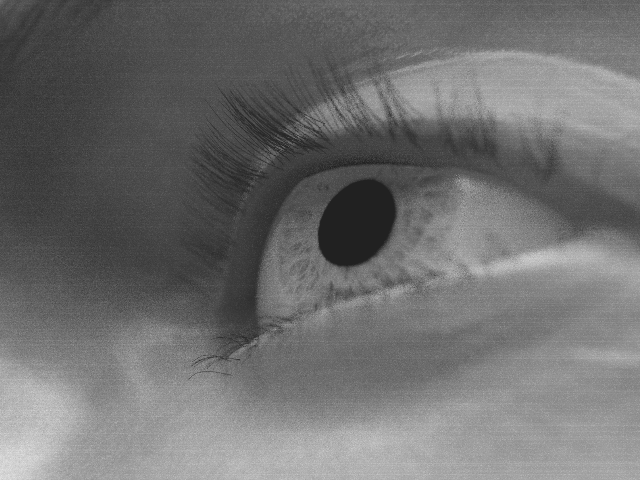}}
	 \put(95, 21){\includegraphics[width=0.15\columnwidth]{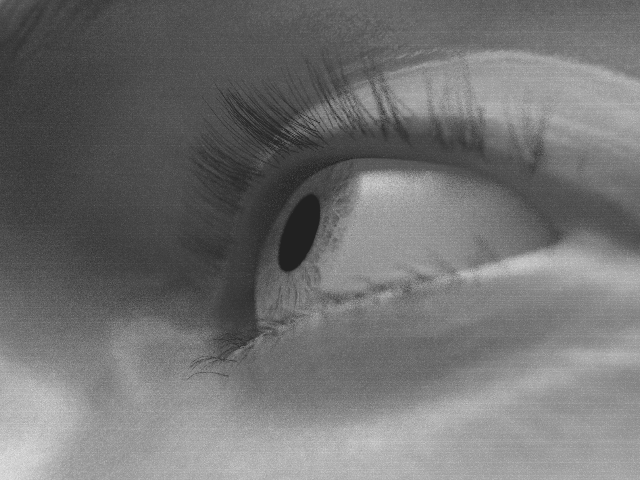}}
	 \put(140,-3){\includegraphics[width=.33\textwidth]{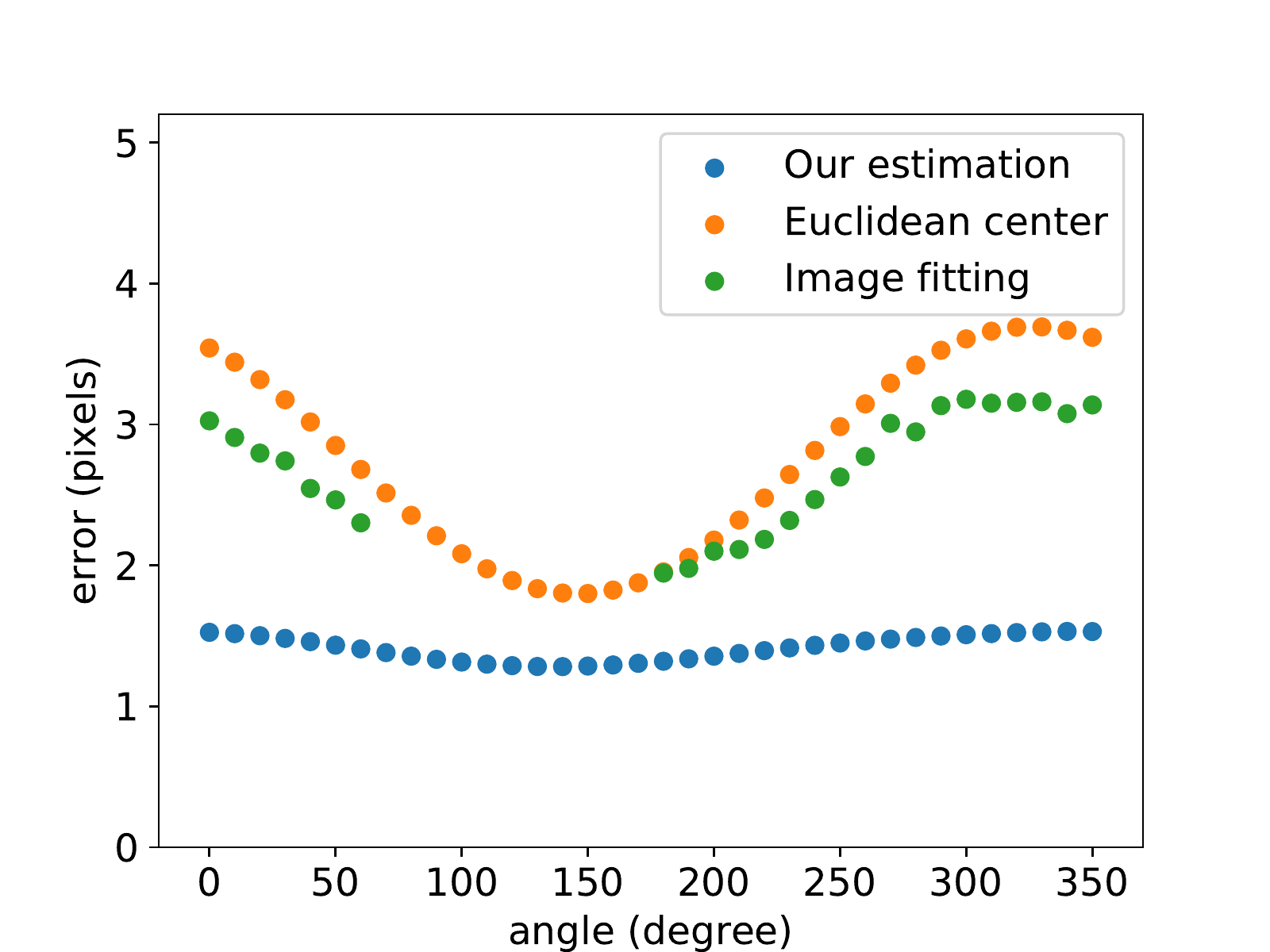}}
	 \put(150, 31){\includegraphics[width=0.15\columnwidth]{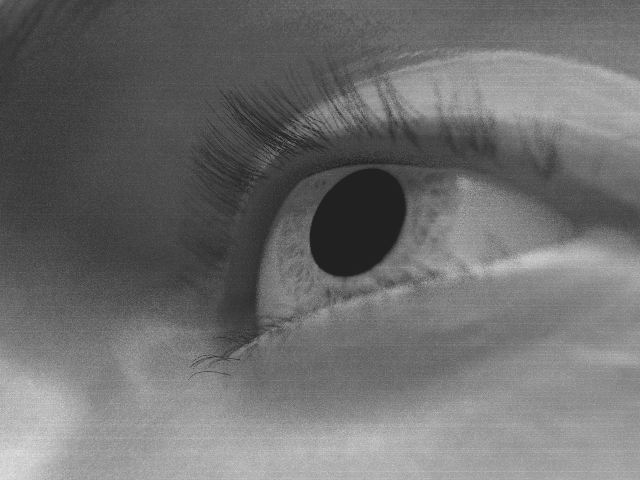}}
	 \put(165, 21){\includegraphics[width=0.15\columnwidth]{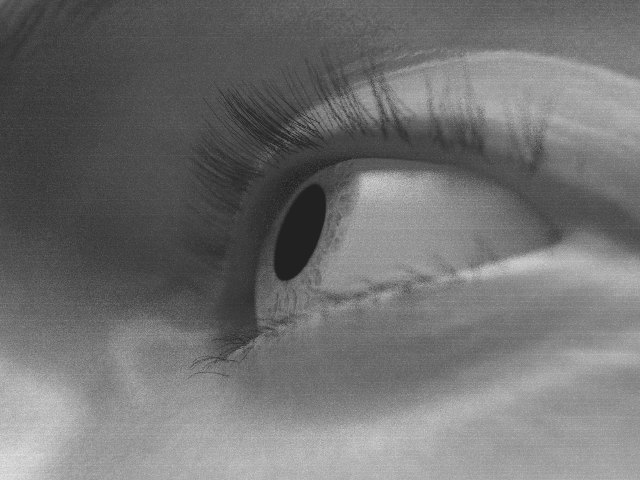}}
    \end{picture}
\caption{Estimation errors with various pupil size. Camera is placed at a fixed position in front of the eye while fixation targets are evenly sampled from a circle in front. Pupil size increases from left to right and embedded eye images show examples when the eyes are rotated to the left and right. $x$ axis corresponds to rotation angle of the eyes in degree and $y$ axis is the estimation error measured in pixels.}
\label{fig:ps}
\end{figure*}

As we see from previous test, the estimation accuracy of the Euclidean ellipse center is correlated to the pupil size in the image plane. In this second experiment, we experiment with three different pupil sizes and compare the estimations when the eyes look at different targets. We evenly place 36 targets on a circle that lies perpendicular to the ground. Figure~\ref{fig:ps} shows the results where estimation error measured in pixel is the y-axis and x-axis corresponds to the rotation angle of the eye measured in degree. As expected, large pupil size leads to large estimation error. All estimation methods get worse but our method consistently provides the most accurate estimation. Once again, image based ellipse fitting method fails when the eyes are tilted to some extend as shown by the second image in each plot. 

\section{Discussion and Future Work}

We introduce a simple algorithm to robustly estimate the true center position from a pair of ellipses projected from a pair of concentric circles. We apply our method in the context of eye tracking and use it to find the true pupil center in the image plane. Evaluation based on synthetic data shows promising results, especially comparing to the estimated center of the fitted ellipse. Even though we did not evaluate the performance of the method on the estimation of pupil size, we believe it is possible to estimate the pupil size using the radii ratio. 

Despite the fact that our method can accurately estimate the true pupil center under projective distortion, it does not consider refraction, making it unsuitable for real-world eye tracking applications. However, note that our formulation allows us to estimate the radii ratio of two concentric circles, the pupil and the iris in this case, and the iris boundary is much less affected by the refraction at the corner. Theoretically we could use the estimated ratio to find the exact position of the pupil center as well as the iris center in the camera image. In other words, this would allow us to implicitly model the corner refraction and find the true pupil center under both distortions in the real scenarios. 

Beyond the scope of eye tracking, concentric circles pattern has been commonly used as fiducial markers in computer vision related tasks~\cite{7780436, DeGol_2017_ICCV}. Since detection accuracy and speed are crucial for fiducial marker based real-time application, our method could provide another option for fiducial marker based tracking. We are eager to investigate in this direction in future work.
 



\bibliographystyle{abbrv-doi}

\bibliography{main}

\begin{thebibliography}{10}

\bibitem{Mora2014}
K.~Alberto Funes~Mora and J.-M. Odobez.
\newblock Geometric generative gaze estimation (g3e) for remote rgb-d cameras.
\newblock In {\em The IEEE Conference on Computer Vision and Pattern
  Recognition (CVPR)}, June 2014.

\bibitem{atchison2000optics}
D.~A. Atchison, G.~Smith, and G.~Smith.
\newblock {\em Optics of the human eye}.
\newblock Butterworth-Heinemann Oxford, 2000.

\bibitem{barsingerhorn2017optics}
A.~Barsingerhorn, F.~Boonstra, and H.~Goossens.
\newblock Optics of the human cornea influence the accuracy of stereo
  eye-tracking methods: a simulation study.
\newblock {\em Biomedical optics express}, 8(2):712--725, 2017.

\bibitem{7780436}
L.~Calvet, P.~Gurdjos, C.~Griwodz, and S.~Gasparini.
\newblock Detection and accurate localization of circular fiducials under
  highly challenging conditions.
\newblock In {\em 2016 IEEE Conference on Computer Vision and Pattern
  Recognition (CVPR)}, pp. 562--570, June 2016. doi: {{%
10\hspace{.1pt}\discretionary{.}{%
}{.}\hspace{.4pt}1109\discretionary{/}{%
}{/}CVPR\hspace{.1pt}\discretionary{.}{%
}{.}\hspace{.4pt}2016\hspace{.1pt}\discretionary{.}{%
}{.}\hspace{.4pt}67}}


\bibitem{DeGol_2017_ICCV}
J.~DeGol, T.~Bretl, and D.~Hoiem.
\newblock Chromatag: A colored marker and fast detection algorithm.
\newblock In {\em The IEEE International Conference on Computer Vision (ICCV)},
  Oct 2017.

\bibitem{Dierkes2018}
K.~Dierkes, M.~Kassner, and A.~Bulling.
\newblock A novel approach to single camera, glint-free 3d eye model fitting
  including corneal refraction.
\newblock In {\em Proceedings of the 2018 ACM Symposium on Eye Tracking
  Research \& Applications}, ETRA '18, pp. 9:1--9:9. ACM, New York, NY, USA,
  2018. doi: {{%
10\hspace{.1pt}\discretionary{.}{%
}{.}\hspace{.4pt}1145\discretionary{/}{%
}{/}3204493\hspace{.1pt}\discretionary{.}{%
}{.}\hspace{.4pt}3204525}}


\bibitem{duchowski2007eye}
A.~T. Duchowski.
\newblock Eye tracking methodology.
\newblock {\em Theory and practice}, 328, 2007.

\bibitem{6630694}
J.~Faigl, T.~Krajník, J.~Chudoba, L.~Přeučil, and M.~Saska.
\newblock Low-cost embedded system for relative localization in robotic swarms.
\newblock In {\em 2013 IEEE International Conference on Robotics and
  Automation}, pp. 993--998, May 2013. doi: {{%
10\hspace{.1pt}\discretionary{.}{%
}{.}\hspace{.4pt}1109\discretionary{/}{%
}{/}ICRA\hspace{.1pt}\discretionary{.}{%
}{.}\hspace{.4pt}2013\hspace{.1pt}\discretionary{.}{%
}{.}\hspace{.4pt}6630694}}


\bibitem{Fedtke10}
C.~Fedtke, F.~Manns, and A.~Ho.
\newblock The entrance pupil of the human eye: a three-dimensional model as a
  function of viewing angle.
\newblock {\em Opt. Express}, 18(21):22364--22376, Oct 2010. doi: {{%
10\hspace{.1pt}\discretionary{.}{%
}{.}\hspace{.4pt}1364\discretionary{/}{%
}{/}OE\hspace{.1pt}\discretionary{.}{%
}{.}\hspace{.4pt}18\hspace{.1pt}\discretionary{.}{%
}{.}\hspace{.4pt}022364}}


\bibitem{Fuhl2016}
W.~Fuhl, T.~C. Santini, T.~K\"{u}bler, and E.~Kasneci.
\newblock Else: Ellipse selection for robust pupil detection in real-world
  environments.
\newblock In {\em Proceedings of the Ninth Biennial ACM Symposium on Eye
  Tracking Research \& Applications}, ETRA '16, pp. 123--130. ACM, New York,
  NY, USA, 2016. doi: {{%
10\hspace{.1pt}\discretionary{.}{%
}{.}\hspace{.4pt}1145\discretionary{/}{%
}{/}2857491\hspace{.1pt}\discretionary{.}{%
}{.}\hspace{.4pt}2857505}}


\bibitem{1634506}
E.~D. Guestrin and M.~Eizenman.
\newblock General theory of remote gaze estimation using the pupil center and
  corneal reflections.
\newblock {\em IEEE Transactions on Biomedical Engineering}, 53(6):1124--1133,
  June 2006. doi: {{%
10\hspace{.1pt}\discretionary{.}{%
}{.}\hspace{.4pt}1109\discretionary{/}{%
}{/}TBME\hspace{.1pt}\discretionary{.}{%
}{.}\hspace{.4pt}2005\hspace{.1pt}\discretionary{.}{%
}{.}\hspace{.4pt}863952}}


\bibitem{Yeong2013}
S.~Y. Gwon, C.~W. Cho, H.~C. Lee, W.~O. Lee, and K.~R. Park.
\newblock Robust eye and pupil detection method for gaze tracking.
\newblock {\em International Journal of Advanced Robotic Systems}, 10(2):98,
  2013. doi: {{%
10\hspace{.1pt}\discretionary{.}{%
}{.}\hspace{.4pt}5772\discretionary{/}{%
}{/}55520}}


\bibitem{holmqvist2011eye}
K.~Holmqvist, M.~Nystr{\"o}m, R.~Andersson, R.~Dewhurst, H.~Jarodzka, and
  J.~Van~de Weijer.
\newblock {\em Eye tracking: A comprehensive guide to methods and measures}.
\newblock OUP Oxford, 2011.

\bibitem{HOOGE20191}
I.~T. Hooge, R.~S. Hessels, and M.~Nyström.
\newblock Do pupil-based binocular video eye trackers reliably measure
  vergence?
\newblock {\em Vision Research}, 156:1 -- 9, 2019. doi: {{%
10\hspace{.1pt}\discretionary{.}{%
}{.}\hspace{.4pt}1016\discretionary{/}{%
}{/}j\hspace{.1pt}\discretionary{.}{%
}{.}\hspace{.4pt}visres\hspace{.1pt}\discretionary{.}{%
}{.}\hspace{.4pt}2019\hspace{.1pt}\discretionary{.}{%
}{.}\hspace{.4pt}01\hspace{.1pt}\discretionary{.}{%
}{.}\hspace{.4pt}004}}


\bibitem{1541275}
G.~Jiang and L.~Quan.
\newblock Detection of concentric circles for camera calibration.
\newblock In {\em Tenth IEEE International Conference on Computer Vision
  (ICCV'05) Volume 1}, vol.~1, pp. 333--340 Vol. 1, Oct 2005. doi: {{%
10\hspace{.1pt}\discretionary{.}{%
}{.}\hspace{.4pt}1109\discretionary{/}{%
}{/}ICCV\hspace{.1pt}\discretionary{.}{%
}{.}\hspace{.4pt}2005\hspace{.1pt}\discretionary{.}{%
}{.}\hspace{.4pt}73}}


\bibitem{Kassner2014}
M.~Kassner, W.~Patera, and A.~Bulling.
\newblock Pupil: An open source platform for pervasive eye tracking and mobile
  gaze-based interaction.
\newblock In {\em Proceedings of the 2014 ACM International Joint Conference on
  Pervasive and Ubiquitous Computing: Adjunct Publication}, UbiComp '14
  Adjunct, pp. 1151--1160. ACM, New York, NY, USA, 2014. doi: {{%
10\hspace{.1pt}\discretionary{.}{%
}{.}\hspace{.4pt}1145\discretionary{/}{%
}{/}2638728\hspace{.1pt}\discretionary{.}{%
}{.}\hspace{.4pt}2641695}}


\bibitem{Palinko2010}
O.~Palinko, A.~L. Kun, A.~Shyrokov, and P.~Heeman.
\newblock Estimating cognitive load using remote eye tracking in a driving
  simulator.
\newblock In {\em Proceedings of the 2010 Symposium on Eye-Tracking Research
  \&\#38; Applications}, ETRA '10, pp. 141--144. ACM, New York, NY, USA, 2010.
  doi: {{%
10\hspace{.1pt}\discretionary{.}{%
}{.}\hspace{.4pt}1145\discretionary{/}{%
}{/}1743666\hspace{.1pt}\discretionary{.}{%
}{.}\hspace{.4pt}1743701}}


\bibitem{Palmero2018RecurrentCF}
C.~Palmero, J.~Selva, M.~A. Bagheri, and S.~Escalera.
\newblock Recurrent cnn for 3d gaze estimation using appearance and shape cues.
\newblock In {\em BMVC}, 2018.

\bibitem{Park2018}
S.~Park, X.~Zhang, A.~Bulling, and O.~Hilliges.
\newblock Learning to find eye region landmarks for remote gaze estimation in
  unconstrained settings.
\newblock In {\em Proceedings of the 2018 ACM Symposium on Eye Tracking
  Research \& Applications}, ETRA '18, pp. 21:1--21:10. ACM, New York, NY, USA,
  2018. doi: {{%
10\hspace{.1pt}\discretionary{.}{%
}{.}\hspace{.4pt}1145\discretionary{/}{%
}{/}3204493\hspace{.1pt}\discretionary{.}{%
}{.}\hspace{.4pt}3204545}}


\bibitem{Swirski2012}
L.~\'{S}wirski, A.~Bulling, and N.~Dodgson.
\newblock Robust real-time pupil tracking in highly off-axis images.
\newblock In {\em Proceedings of the Symposium on Eye Tracking Research and
  Applications}, ETRA '12, pp. 173--176. ACM, New York, NY, USA, 2012. doi: {{%
10\hspace{.1pt}\discretionary{.}{%
}{.}\hspace{.4pt}1145\discretionary{/}{%
}{/}2168556\hspace{.1pt}\discretionary{.}{%
}{.}\hspace{.4pt}2168585}}


\bibitem{Swirski2014}
L.~\'{S}wirski and N.~Dodgson.
\newblock Rendering synthetic ground truth images for eye tracker evaluation.
\newblock In {\em Proceedings of the Symposium on Eye Tracking Research and
  Applications}, ETRA '14, pp. 219--222. ACM, New York, NY, USA, 2014. doi: {{%
10\hspace{.1pt}\discretionary{.}{%
}{.}\hspace{.4pt}1145\discretionary{/}{%
}{/}2578153\hspace{.1pt}\discretionary{.}{%
}{.}\hspace{.4pt}2578188}}


\bibitem{Swirski2013}
L.~\'Swirski and N.~A. Dodgson.
\newblock A fully-automatic, temporal approach to single camera, glint-free 3d
  eye model fitting [abstract].
\newblock In {\em Proceedings of ECEM 2013}, Aug. 2013.

\bibitem{szczepanski2014pupil}
A.~Szczepa{\'n}ski, K.~Misztal, and K.~Saeed.
\newblock Pupil and iris detection algorithm for near-infrared capture devices.
\newblock In {\em IFIP International Conference on Computer Information Systems
  and Industrial Management}, pp. 141--150. Springer, 2014.

\bibitem{UKAI2008106}
K.~Ukai and P.~A. Howarth.
\newblock Visual fatigue caused by viewing stereoscopic motion images:
  Background, theories, and observations.
\newblock {\em Displays}, 29(2):106 -- 116, 2008.
\newblock Health and Safety Aspects of Visual Displays. doi: {{%
10\hspace{.1pt}\discretionary{.}{%
}{.}\hspace{.4pt}1016\discretionary{/}{%
}{/}j\hspace{.1pt}\discretionary{.}{%
}{.}\hspace{.4pt}displa\hspace{.1pt}\discretionary{.}{%
}{.}\hspace{.4pt}2007\hspace{.1pt}\discretionary{.}{%
}{.}\hspace{.4pt}09\hspace{.1pt}\discretionary{.}{%
}{.}\hspace{.4pt}004}}


\bibitem{4564194}
A.~Villanueva and R.~Cabeza.
\newblock Evaluation of corneal refraction in a model of a gaze tracking
  system.
\newblock {\em IEEE Transactions on Biomedical Engineering}, 55(12):2812--2822,
  Dec 2008. doi: {{%
10\hspace{.1pt}\discretionary{.}{%
}{.}\hspace{.4pt}1109\discretionary{/}{%
}{/}TBME\hspace{.1pt}\discretionary{.}{%
}{.}\hspace{.4pt}2008\hspace{.1pt}\discretionary{.}{%
}{.}\hspace{.4pt}2002152}}


\bibitem{Zhang2019}
X.~Zhang, Y.~Sugano, M.~Fritz, and A.~Bulling.
\newblock Mpiigaze: Real-world dataset and deep appearance-based gaze
  estimation.
\newblock {\em IEEE Transactions on Pattern Analysis and Machine Intelligence},
  41(1):162--175, Jan 2019. doi: {{%
10\hspace{.1pt}\discretionary{.}{%
}{.}\hspace{.4pt}1109\discretionary{/}{%
}{/}TPAMI\hspace{.1pt}\discretionary{.}{%
}{.}\hspace{.4pt}2017\hspace{.1pt}\discretionary{.}{%
}{.}\hspace{.4pt}2778103}}


\end{thebibliography}
\end{document}